\documentclass[10pt,twocolumn,letterpaper]{article}

\usepackage{iccv}
\usepackage{times}
\usepackage{epsfig}
\usepackage{graphicx}
\usepackage{amsmath}
\usepackage{amssymb}

\graphicspath{{./res/figs/}}
\usepackage{booktabs}
\usepackage{multirow}
\usepackage{makecell}
\usepackage{xcolor}
\usepackage{bbding}
\usepackage{pifont}

\usepackage{float}

\usepackage[pagebackref=true,breaklinks=true,letterpaper=true,colorlinks,bookmarks=false]{hyperref}

\iccvfinalcopy 

\ificcvfinal\fi

\begin{document}

\title{MVP-SEG: Multi-View Prompt Learning for Open-Vocabulary Semantic Segmentation}

\author{Jie Guo\textsuperscript{1}\thanks{These authors contributed equally.} \quad Qimeng Wang\textsuperscript{2}\footnotemark[1] \quad Yan Gao\textsuperscript{2} \quad Xiaolong Jiang\textsuperscript{2}\\
Xu Tang\textsuperscript{2} \quad Yao Hu\textsuperscript{2} \quad Baochang Zhang\textsuperscript{1,$\dag$}\\
\\
\textsuperscript{1}Beihang University \quad \textsuperscript{2}Xiaohongshu Inc\\
}

\twocolumn[{
\maketitle
\begin{figure}[H]
\hsize=\textwidth
\centering
\includegraphics[width=\textwidth]{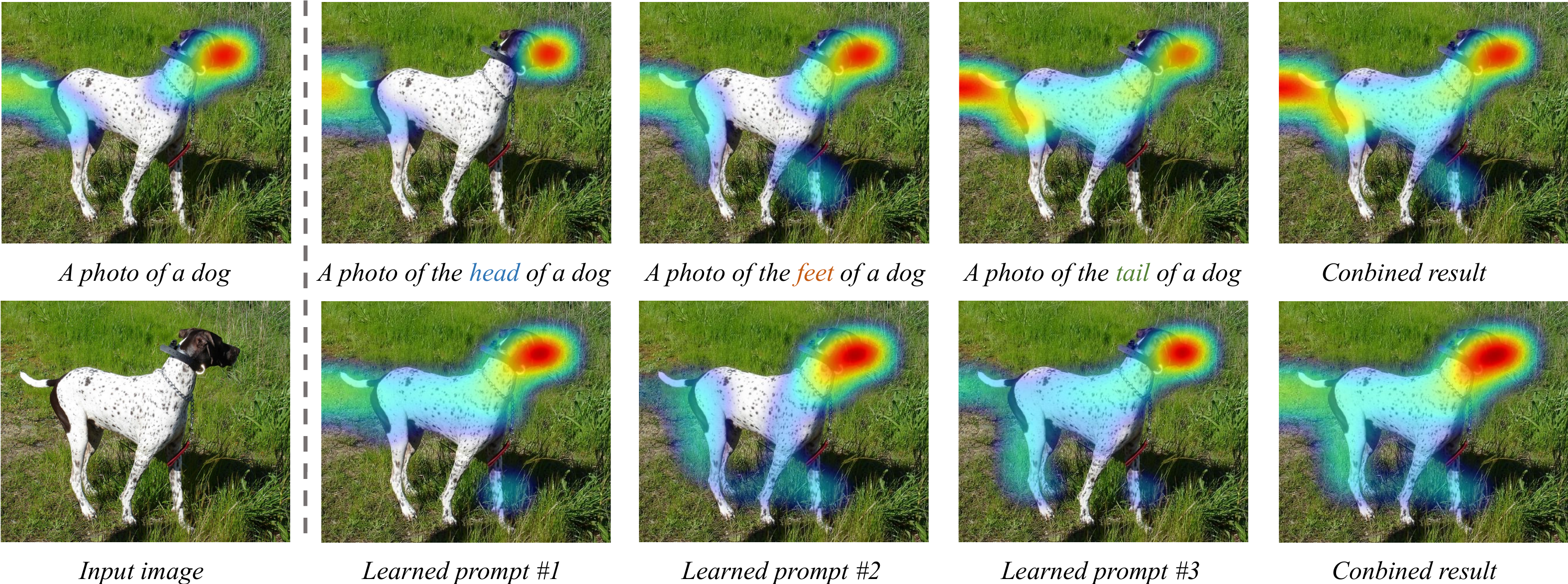}
\caption{The illustrated motivation of multi-view learnable prompts. The first row demonstrates that even handcrafted, part-based multi-view prompts collaboratively show better localization capability \it{w.r.t.} the object, such that attention responses in the right-most picture are more prominent than that in the left-most picture (which results from the original single handcrafted prompt). The second row exhibits that multi-view learnable prompts effectively capture the part-based characteristics and provide better combined result. Notedly, multi-view learnable prompts are flexible to accommodate open-world objects without being fixed to specific concepts such as "head" or "tail".}
\label{fig:fig1}
\end{figure}
}]
\let\thefootnote\relax\footnotetext{$\star$ Equal contribution. $\dag$ Corresponding author.}

\begin{abstract}
CLIP (Contrastive Language-Image Pretraining) is well-developed for open-vocabulary zero-shot image-level recognition, while its applications in pixel-level tasks are less investigated, where most efforts directly adopt CLIP features without deliberative adaptations. In this work, we first demonstrate the necessity of image-pixel CLIP feature adaption, then provide Multi-View Prompt learning (MVP-SEG) as an effective solution to achieve image-pixel adaptation and to solve open-vocabulary semantic segmentation. Concretely, MVP-SEG deliberately learns multiple prompts trained by our Orthogonal Constraint Loss (OCLoss), by which each prompt is supervised to exploit CLIP feature on different object parts, and collaborative segmentation masks generated by all prompts promote better segmentation. Moreover, MVP-SEG introduces Global Prompt Refining (GPR) to further eliminate class-wise segmentation noise. Experiments show that the multi-view prompts learned from seen categories have strong generalization to unseen categories, and MVP-SEG+ which combines the knowledge transfer stage significantly outperforms previous methods on several benchmarks. Moreover, qualitative results justify that MVP-SEG does lead to better focus on different local parts.  
\end{abstract}

\section{Introduction}
\label{sec:intro}
Open-vocabulary zero shot semantic segmentation \cite{spnet,zs3net,cagnet,sign,joint,strict} localizes objects from arbitrary classes, either seen or unseen during training time, with pixel-level masks. Compared to traditional segmentors working under closed-vocabulary setting \cite{deeplab,kim2022restr,yuan2020object,jin2021isnet,zegformer,maskclip}, open-vocabulary methods find wider applications in image editing \cite{liu2020open}, view synthesis \cite{qian2022multimodal}, and surveillance \cite{yu2022argus++}, while requiring to integrate zero-shot capability into the system. 

The de facto solution for integrating zero-shot capability into visual systems is via large-scale visual-language pre-training models such as CLIP \cite{clip} and ALIGN \cite{align}. Specifically, CLIP encodes semantic concepts into its parameters through contrastive training on web-scale image-text pairs, becoming a zero-shot knowledge base for downstream tasks. It is worth noting that, contrastive pre-training mainly focuses on capturing image-level concepts. The training text used in CLIP mostly describes global contexts of the images and the encoded image and text embeddings are used in their entirety to compute contrastive loss. Accordingly, CLIP is more suitable for image-level classification \cite{coop, cocoop, learnpromptdist, tipadapter, huang2022unsupervised},  but less-developed in pixel-level segmentation.

Amongst the few pixel-level attempts, MaskCLIP \cite{maskclip} pioneers in adopting CLIP for open-vocabulary segmentation, where pre-trained CLIP text features are directly used as the semantic classifier to perform pixel-level classification without further adaptation. We, however, believe image-to-pixel CLIP adaptation can further improve segmentation performance compared to direct adoption. As shown in Fig~\ref{fig:fig1}, original CLIP prompt features tend to focus only on the most class-discriminative parts of the objects, this phenomenon occurs as CLIP is pre-trained using image-level contrastive loss, and thus is prone to induce incomplete and partial segmentation. Therefore, we advocate the importance of performing image-to-pixel adaptation when CLIP is applied in open-vocabulary segmentation tasks, which is also supported by \cite{zegformer, cha2022zero}, wherein both ImageNet and Vision-Language pre-trained backbones require adaptations employed in box or pixel-level tasks. 

We seek prompt learning to adapt CLIP features from image-to-pixel while keeping zero-shot capability intact. Prompt learning~\cite{huang2022unsupervised,coop} eschews prohibitive cost of CLIP training and adapts CLIP towards down-stream tasks via fine-tuning limited extra prompt parameters. To further customize prompt learning towards effective image-to-pixel adaptation for segmentation, inspired by the part-based representation widely adopted in traditional segmentation methods~\cite{yu2019partnet, eslami2012generative}, we propose to exploit multiple prompts to capture different object parts and collaboratively they can deliver more accurate and complete segmentation. To visually validate this proposal, we manually add different body parts into multiple prompts, e.g. \textit{`` A photo of the \textbf{head} of a \textbf{dog}''} and \textit{`` A photo of the \textbf{tail} of a \textbf{dog}''}. As shown in Fig~\ref{fig:fig1}, different prompts do attend to corresponding parts in the image, and by combining responses of all prompts we achieve better coverage of the object. On top of this, by switching handcrafted prompts with learnable ones, we are not confined to fixed prompt templates and thus generalizing to open-vocabulary classes.

Based on the above discussions, we propose MVP-SEG: Multi-View Prompt learning for open-vocabulary semantic segmentation. In MVP-SEG, we start by utilizing one learnable prompt instead of the default handcrafted prompt for image-to-pixel CLIP adaptation. As shown in Fig.~\ref{fig:num_prompts}, one learnable prompt can already boost segmentation performance and further substantiate the necessity of the image-to-pixel adaptation. On top of this, we extend to multi-view learnable prompts for the part-based representation. In order to ensure each prompt attends to different object parts during segmentation, we introduce the Orthogonal Constraint Loss (OCLoss) which regularizes learned prompts to be mutually orthogonal. Additionally, we design Global Prompt Refining (GPR) module to further improve segmentation performance by eliminating class-wise noise from a global perspective. Extensive experiments reveal that our method surpasses current SOTA, and source code will be released upon acceptance.

In all, we propose three main contributions:
\begin{itemize}
    \item We demonstrate that image-to-pixel adaptation is important for adopting exploiting CLIP’s zero-shot capability in open-vocabulary semantic segmentation, and the proposed MVP-SEG successfully yields such adaptation with favorable performance gains.
    \item  We design the OCLoss to build multi-view learnable prompts attending to different object parts so that collaboratively they yield accurate and complete segmentation. We also introduce GPR to eliminate class-wise segmentation noise.
    \item We conduct extensive experiments on three major benchmarks, MVP-SEG+ which combine MVP-SEG with the commonly used knowledge transfer stage reports state-of-the-art(SOTA) performance on all benchmarks and even surpasses fully-supervised counterparts on Pascal VOC and Pascal Context datasets.
\end{itemize}

\begin{figure*}[t]
\begin{center}
\includegraphics[width=.98\linewidth]{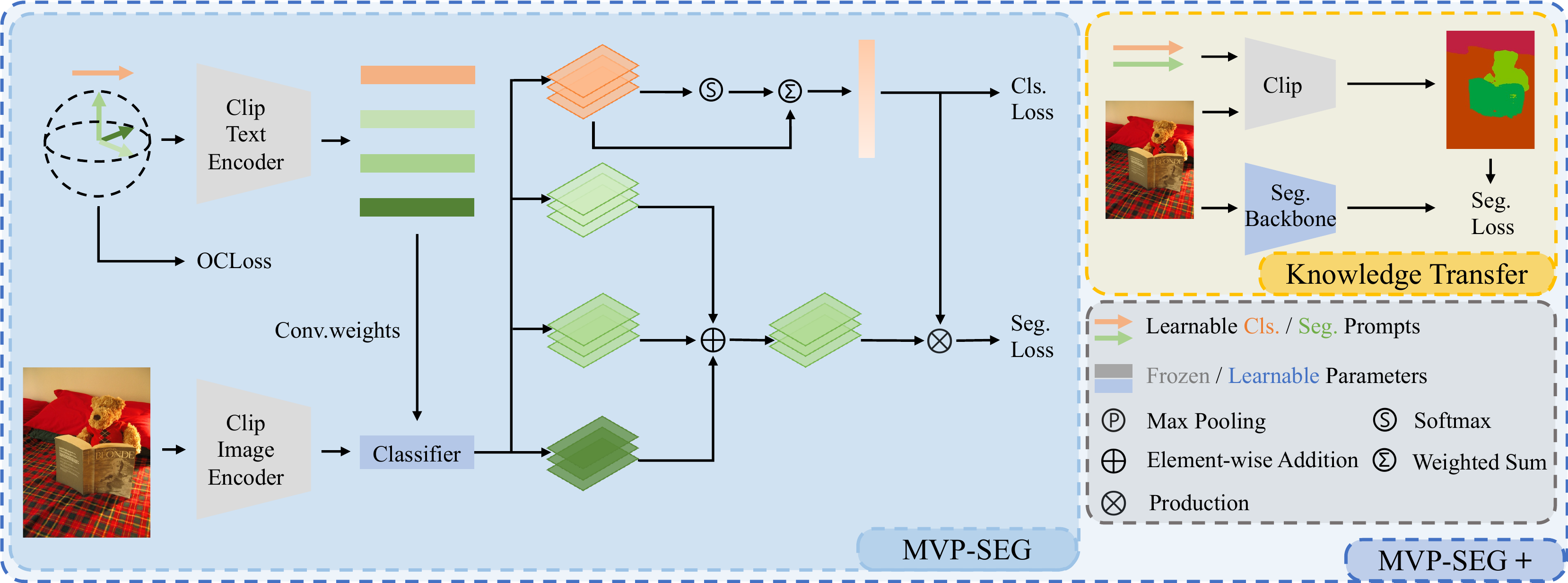}
\end{center}
   \caption{The overview of MVP-SEG and MVP-SEG+. The MVP-SEG+ consists of MVP-SEG and a knowledge transfer stage. In MVP-SEG, we train multi-view prompts by Classification Loss, Segmentation Loss and Orthogonal Constraint Loss. The CLIP text and image encoders are fixed. In the knowledge transfer stage, the segmentation backbone is trained by the pseudo labels generated by MVP-SEG.}
\label{fig:architecture}
\end{figure*}

\section{Related Work}
\subsection{Vision-language Models}
Vision-language (VL) models pretrained with web-scale image-text pairs infuse open-world knowledge into aligned textual and visual features, upon which zero-shot visual recognition\cite{yue2021counterfactual} and generation \cite{dalle, blip2} bloom in recent years. Recognition-purpose VL models can be categorized as two-stream \cite{clip, align, vilbert} or one-stream \cite{visualbert, vlbert, uniter} depending on whether multimodal inputs are processed separately or all together. Amongst, CLIP \cite{clip} is more popular in downstream applications and it is contrastively trained using 400 million image-text pairs. ALIGN \cite{align} adopts even larger dataset with 1.8 billion pairs but with considerable noises. More recently, CoCa \cite{coca} and Beit-V3 \cite{beitv3} further highlight the superiority of VL pretrained features by dominating most down-stream vision as well as multimodal tasks. For zero-shot generation, DALL-E \cite{dalle} and BLIP-v2 \cite{blip2} have shown outstanding generation quality in executing text-guided image generation. In this work we build upon the zero-shot capability of CLIP to realize semantic segmentation in open vocabulary.  

\subsection{Pre-Training and Self-Training}
Pre-training and self-training are two supervision schemes optimizing models towards final goal with additional labels. Pre-training denotes supervising models on large-scale datasets (vision data such as ImageNet \cite{imagenet} and JFT \cite{jft}, vision-language data such as Conceptual12M \cite{conceptual}, VisualGenome \cite{visualgnome}, and VLIGN \cite{align}) and then applied down-stream tasks via task-specific finetuning or prompting. Self-training is semi-supervised such that fully-supervised models are used to produce pseudo labels on unlabeled datasets, then these models are fine-tuned with both labeled and pseudo-labeled data \cite{zegformer}. Self-training has been practiced in image classification \cite{xie2020self}, object detection \cite{yang2021interactive}, as well as action recognition \cite{dave2022spact}, and has also been found effective in semantic segmentation \cite{maskclip}. Our MVP-SEG is built with both pre-training and self-training. 

\subsection{Zero-Shot Segmentation}
Zero-shot segmentation performs pixel-level classification covering unseen classes during training. SPNet\cite{spnet} and ZS3Net \cite{zs3net} are the representatives of discriminative \cite{joint,strict,spnet} and generative \cite{zs3net,cagnet,conterfactual} lines of work, the former projects visual embeddings towards semantic embeddings, while the later generates pixel-level features for the unseen using semantic embeddings. Following ZS3Net, fruitful generative methods are proposed \cite{uncertainty, consistent, cagnet, zegformer, conterfactual, cagnet, cap2seg, zerois} to mind the object-to-pixel feature gap considering structural consistency and uncertainty. Besides, self-training is widely-adopted in zero-shot segmentation \cite{zs3net, strict} to boost performance, and CLIP is also exploited to bridge the text-to-visual feature space \cite{denseclip, cpt, maskclip}.

\subsection{Prompt Learning}
The idea of prompting originates from NLP (Nature Language Processing) and has been exploited for transferring VL-pretrained models to down-stream task in forms like “a photo of a [CLS]”. Via prompting, one can eschew tuning huge VL models but to use it as a fixed knowledge base, wherein only task-related information is elicited. Nonetheless, finding the optimal prompt is not trivial by hand, thus prompt learning \cite{coop} is proposed to automate this process with limited labeled data. CoOp \cite{coop} introduces continuous prompt learning such that a set of continuous vectors are end-to-end optimized via down-stream supervision~\cite{lester2021power,li2021prefix}, and CoOp applies learnable prompts on the text encoder of CLIP to replace sub-optimal hand-crafted templates. Our MVP-SEG further extends CoOp to a multi-view paradigm, founded on the fact that part-based representation is beneficial for segmentation task.

\section{Method}
In this section, we first give a brief introduction to the open-vocabulary semantic segmentation task, then we introduce our proposed MVP-SEG and MVP-SEG+ in details.


\subsection{Problem Definition}
Semantic segmentation performs pixel-level classification to localize objects from different classes in the input image. Under traditional close-vocabulary settings, target classes form a finite set, and all classes are present in both training and test sets. Open-vocabulary semantic segmentation aims to generalize beyond finite base classes and segments objects from novel classes unseen during training. In order to quantitatively evaluate segmentation performance under this open-vocabulary setting, object classes in existing segmentation datasets are divided as seen and unseen subsets \cite{maskclip, strict}. Specifically, models are trained on labeled seen classes and evaluated on both seen and unseen classes. Additionally, we also visualize segmentation results on arbitrary rare classes as in Fig.~\ref{fig:arbitary} to further validate our open-vocabulary capability.

\subsection{MVP-SEG}
The architecture of MVP-SEG is 
illustrated in Fig.~\ref{fig:architecture}. Note that MVP-SEG learns multiple prompts by training only on seen classes. 

\noindent \textbf{Multi-View Prompt Learning}.
In this stage, we adopt a two-stream network architecture separately employing an image and a text encoder. We modify a fixed CLIP image encoder as our vision encoder. In specifics, the original CLIP vision encoder adopts a transformer-style attention pooling layer $AttnPool$ on the last feature map $X$ to get the representation vector of the input image. $X$ is projected into $Q$,$K$,$V$ by 3 linear layers $Proj_{q}$, $Proj_{k}$ and $Proj_{v}$ respectively, then $AttnPool$ is performed on $Q$,$K$,$V$ to get the final representation vector as:
\begin{equation}
    AttnPool(Q,K,V) = Proj_{c}(\sum_{i}softmax(\frac{(\bar{q}k_{i}^{T})}{T})v_i))
\end{equation}
where $\bar{q}$ is the spatial-wise averaged $Q$. $k_{i}$ and $v_{i}$ are the features of $K$ and $V$ at spatial location $i$. $T$ is a constant scaling factor.
Following MaskCLIP ~\cite{maskclip}, we remove the query and key embedding layers $Proj_{q}$ and $Proj_{k}$, then apply $Proj_{v}$ and $Proj_{c}$ on feature map $X$ to obtain the final image feature map $F$, where 
\begin{equation}
    F= Proj_{c}(Proj_{v}(X))
\end{equation}
In practice, $Proj_{c}$ and $Proj_{v}$ are implemented as $1\times 1$ convolution layers.

For the text encoder, we use a fixed CLIP text encoder to ensure feature alignment with the visual counterpart. For each class, we combine the class name with learnable prompts to form the inputs for the text encoder. We first initialize k+1 prompts $P$,
where $P=\{p_0,p_1,p_2,...p_k\}$. $p_0$ is the global classification prompt and $p_1,p_2,..p_k$ represent k segmentation prompts. Each prompt $p_i$ contain 32 tokens, where $p_{i}\in \mathbb{R}^{32\times 512}$. For class $c$, we concatenate each prompt with the class name as:
\begin{equation}
s_{c}^{i} = \textit{CONCAT}(p_i,\textit{W}(cls_{c}))
\end{equation}

\noindent here $\textit{W}$ denotes the word embedding function that maps $cls_{c}$ to word embedding vectors. We feed each sentence $s_c^{i}$ to the CLIP text encoder and get the text representation vector $t_c^i$, where $t_c^0$ and $t_c^i (i>0)$ indicates global classification vector and  segmentation vector of class $c$.

Each representation vector $t_{c}^i$ is used as a classifier to perform pixel-level classification on the feature map $F$ to get the mask map $m_{c}^i$, where $m_{c}^0$ is the global classification mask and $m_c^1,m_c^2...m_c^{k}$ are the segmentation masks.

The fused segmentation result $m^{f}$ is computed by summing all segmentation masks as:
\begin{equation}
m^{f} = softmax(\tau_1 \sum_{i=1} ^{k} m^i )
\end{equation}
where $\tau_1$ is a learnable scalar parameter.

\begin{table*}[t]
\centering
\scalebox{0.9}{
\begin{tabular}{c c c c c c c c c c c c c }
\toprule
\multirow{2}{*}{Method}&\multirow{2}{*}{OCLoss}&\multirow{2}{*}{GPR}&\multicolumn{4}{c}{COCO Stuff}&\multicolumn{4}{c}{PASCAL VOC}\\

\cmidrule(lr){4-7}
\cmidrule(lr){8-11}

&&& mIoU(U) &  mIoU(S) & mIoU  & hIoU & mIoU(U) &  mIoU(S) & mIoU  & hIoU\\
\midrule

Baseline         &-&-                   &12.2 &10.0 &10.2 &11.0 &40.2 &41.9 &41.5 &41.1 \\
Single Prompt & \ding{55} &\ding{55}    &15.8 &15.1 &15.2 &15.5 &41.8 &46.9 &45.6 &44.2 \\
Multiple Prompt & \ding{55} &\ding{55}  &16.4 &16.0 &16.0 &16.2 &45.8 &48.4 &47.7 &47.0  \\
Multiple Prompt  & \ding{51} &\ding{55} &19.9 &16.1 &16.4 &17.9 &45.9 &51.1 &49.8 &48.3 \\
Multiple Prompt  & \ding{51} &\ding{51} &\textbf{20.6} &\textbf{17.7} &\textbf{18.0} &\textbf{19.2} &\textbf{52.9} &\textbf{53.4} &\textbf{53.2} &\textbf{53.1} \\

\bottomrule
\end{tabular}}
\caption{Ablation study of results of MVP-SEG on COCO Stuff and PASCAL VOC dataset, we adopt MaskCLIP~\cite{maskclip} as the baseline model for comparison. Note that the number of multiple prompts without GPR is 3.}
\label{tab:ablation}
\end{table*}

In order to ensure different prompts result in masks attending to varied object parts and collectively they provide better segmentation, we design the Orthogonal Constraint Loss (OCLoss) to supervise prompt learning. For each segmentation prompt $p_i \in \mathbb{R}^{32\times 512}$,  we first average the token dimension to get its average vector $p_i^{\prime}$, where $p_i^{\prime} \in \mathbb{R}^{1\times 512}$, then applied the OCLoss which is formulated as: 
\begin{equation}
L_{OC} = \sum_{i=1}^{k}\sum_{j=i+1}^{k} \frac{|p_i^{\prime}  \cdot p_j^{\prime}|}{ \Vert p_i^{\prime} \Vert   \Vert p_j^{\prime} \Vert }
\end{equation}

\noindent \textbf{Global-Prompt Refinement}.To incorporate CLIP's strong capability in zero-shot image classification into segmentation, we introduce Global-Prompt Refinement(GPR) module which uses a global classification prompt to obtain image classification scores and refine the segmentation mask by eliminating class-wise noises. The image-level classification score $g_c$ is obtained by weighting and summing $m_c^0$ as 
\begin{equation}
g_c =  sigmoid(\frac{m_c^0  \cdot softmax(\gamma m_c^0)}{\tau_2})
\end{equation}
where $\tau_2$ is a learnable scalar parameter and $\gamma$ is a constant scale factor. The global classification loss is as:
\begin{equation}
L_{cls} =- \sum_{c} y_clog(g_c)
\end{equation}
here $y_c$ equals 1 if category c exists in this image, and else 0. The classification prompt tends to focus on the most class-discriminative object parts for better classification. More visualizations can be found in the Appendix.

The final segmentation mask $m_c$ is obtained by multiplying the global classification score with fused segmentation masks:
\begin{equation}
m_c = m_c^f g_c 
\end{equation}

The segmentation loss is formulated as:
\begin{equation}
L_{seg} =-\sum^{H}\sum^{W}\sum_{c=1}^{C}m_c^*log(m_c)
\end{equation}
where $m_c^*$ is the ground-truth mask of category $c$. $C$ is the number of segmentation classes. $H$ and $W$ denote the height and the width of the input image respectively.  The overall loss function is obtained by summing the above losses as 
\begin{equation}
L = \lambda_1 L_{seg} + \lambda_2 L_{cls} + \lambda_3 L_{OC}
\end{equation}

Note that in the prompt learning stage, the image encoder and text encoder are fixed, only prompt vectors $P$ and temperature $\tau_1$,$\tau_2$ are learnable, so that the image-text feature alignment within CLIP is preserved. In addition, our experiments also demonstrate that prompts trained on limited seen categories generalize well to unseen categories.

\subsection{MVP-SEG+}
We term the entire framework in Fig.~\ref{fig:architecture} as MVP-SEG+, consisting of MVP-SEG and knowledge transfer stage.
\noindent \textbf{Knowledge Transfer}.
 The learned prompts in MVP-SEG are used to generate pseudo labels to transfer zero-shot knowledge from CLIP down to a segmentation network.

Following the application of MaskCLIP+\cite{maskclip}, the transfer learning stage contains two steps: pseudo-label training and self-training. At the pseudo-label training step, images are fed into MVP-SEG to generate the pseudo labels for the segmentation model. The classifier of segmentation model is replaced by the multi-view prompts of MVP-SEG to preserve the open-vocabulary capability. In self-training stage, the segmentation model in MVP-SEG+ starts to train itself with self-generated pseudo labels. For more details, please refer to MaskCLIP+~\cite{maskclip}.


\section{Experiments}
To evaluate the efficacy of MVP-SEG and MVP-SEG+, we conduct comprehensive experiments on three widely-adopted benchmarks. In this section, we first analysis the performance and ablation study of MVP-SEG. Then, we compare MVP-SEG+ with the SOTA zero-shot segmentation method to show the effectiveness of our proposed method to adapt CLIP to pixel-level tasks. At last, we show the open-vocabulary ability of our method on unlabeled web images.
\subsection{Datasets}
Three semantic segmentation benchmarks are used in our experiment. PASCAL VOC 2012~\cite{pascal-voc-2012} contains 1426 training images with 20 object classes and 1 background class. PASCAL Context \cite{mottaghi2014role} annotates PASCAL VOC 2010 data with segmentation masks of 10,103 images covering 520 classes, among which 59 common classes are used as foregrounds. COCO Stuff dataset \cite{caesar2018coco} is an extension of the COCO dataset, 164,000 images covering 171 classes are annotated with segmentation masks. We follow the common zero-shot experimental setups as implemented in~\cite{maskclip}. For PASCAL VOC, we ignore the background class, and \textit{potted plant, sheep, sofa, train, tv monitor} are selected as 5 unseen classes while others are used as seen. For PASCAL Context, the background is not ignored and \textit{cow, motorbike, sofa, cat, boat, fence, bird, tv monitor, keyboard, aeroplane} are unseen. For COCO Stuff, \textit{ frisbee, skateboard, cardboard, carrot, scissors, suitcase, giraffe, cow, road, wall concrete, tree, grass, river, clouds, playing field} are used as unseen classes.

\subsection{Evaluation Metrics}
We utilize the \textit{mean intersection-over-union} (mIoU) on seen and unseen classes to evaluate semantic segmentation performance. Additionally, we also report the harmonic mean of seen and unseen mIoUs (hIoU).

\subsection{Implementation Details}
We adopt MMSegmentation\footnote {https://github.com/open-mmlab/mmsegmentation} as our codebase. In the prompt learning stage, we use the encoders from CLIP-ResNet-50\footnote{https://github.com/openai/CLIP} to extract image and text features. We use 1 global classification and 3 multi-view segmentation prompts unless otherwise stated. All learnable prompts are implemented using unified context\cite{coop} with 32 context tokens. The weighting parameters $\lambda_1$, $\lambda_2$, $\lambda_3$ for $L_{seg}$,$L_{cls}$ and $L_{OC}$ are empirically set to 1, 3, 100 respectively. The softmax scale factor $\gamma$ is set to 10. We use SGD optimizer to optimize learnable prompts with learning rate set to 2e-4 and 5e-4 weight decay. We also adopt linear warmup strategy with 1k warmup iters and 1e-3 warmup ratio. The prompt learning step takes 8k iterations with batch size 8. All the experiments are conducted using 4 Tesla A100 GPUs. 

In the knowledge transfer stage, we use the same settings as MaskCLIP~\cite{maskclip} for fair comparison. We choose DeepLabv2~\cite{deeplab} as the segmentation backbone for PASCAL VOC and COCO Stuff and DeepLabv3+ for PASCAL Context. The training schedule is set to 20k/40k/80k for PASCAL VOC/PASCAL Context/COCO Stuff. The first 1/10 training iterations adopt MVP-SEG guided learning and the rest adopts self-training. 
\begin{table}
\footnotesize
\centering
  \begin{tabular}{l | c | c | c|c|c|c|c}
    \toprule
    Method 
    & cat
    & dog
    & horse
    & cow
    & bear
    & giraffe
    & mIoU
    \\
    \midrule
    CLIP         &55.2 &39.6 &39.0 &29.5 &31.8 &57.2 &44.3\\
    HC           &57.5 &40.6 &40.1 &31.2 &33.0 &57.9 &45.3\\
    Ours         &\textbf{63.0} &\textbf{42.2} &\textbf{47.4} &\textbf{33.8} &\textbf{36.8} &\textbf{63.8} &\textbf{51.5}\\
    \midrule
  \end{tabular}
  \caption{Performance comparison on animal related categories between Hand-designed prompts and learnable prompts.} 
  \label{tab:tab5}
\end{table}

\begin{figure}[t]
\begin{center}
\includegraphics[width=1\linewidth]{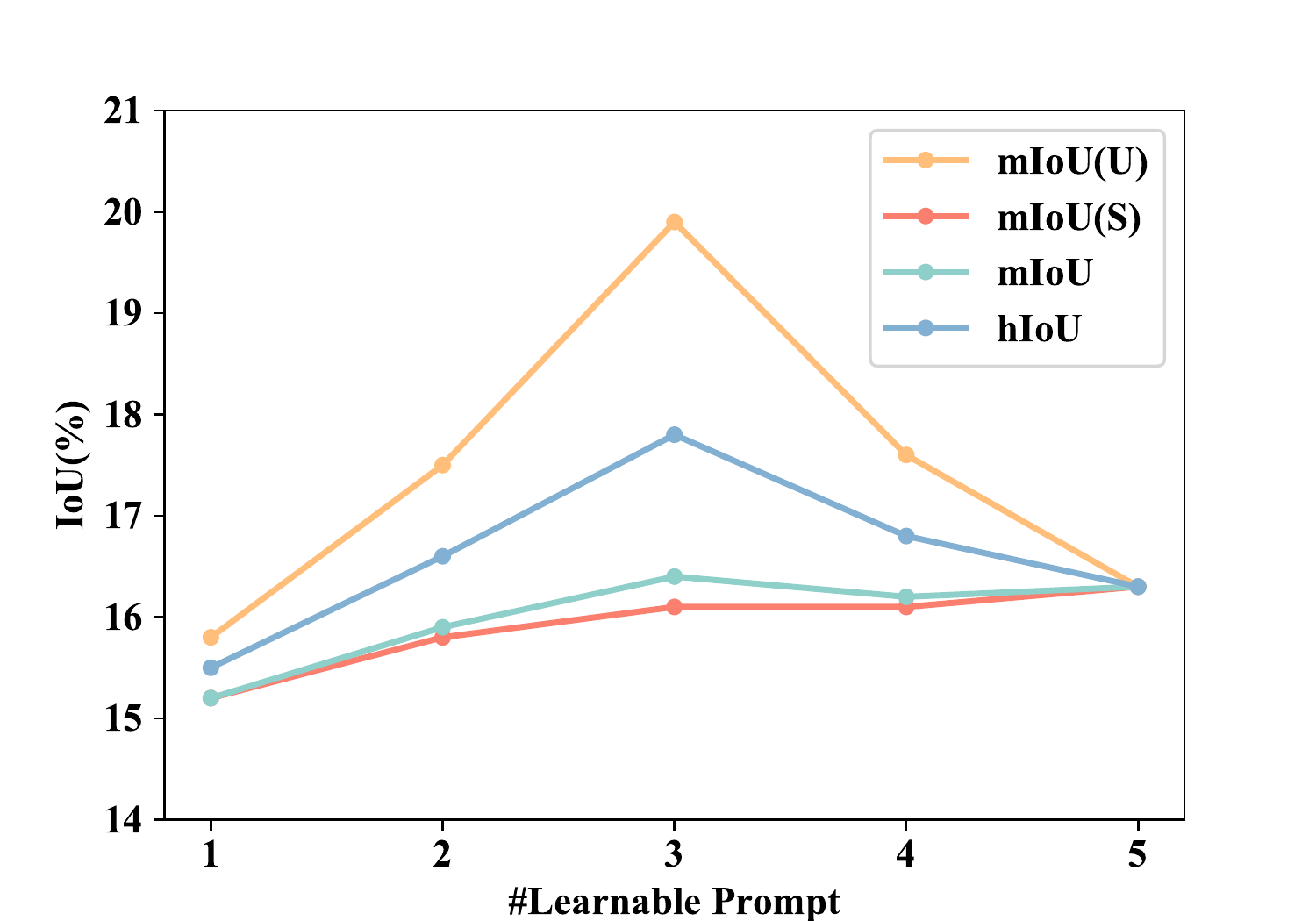}
\end{center}
   \caption{The performance on COCO Stuff dataset with different numbers of segmentation prompts.}
\label{fig:num_prompts}
\end{figure}

\begin{figure}[t]
\begin{center}
\includegraphics[width=.76\linewidth]{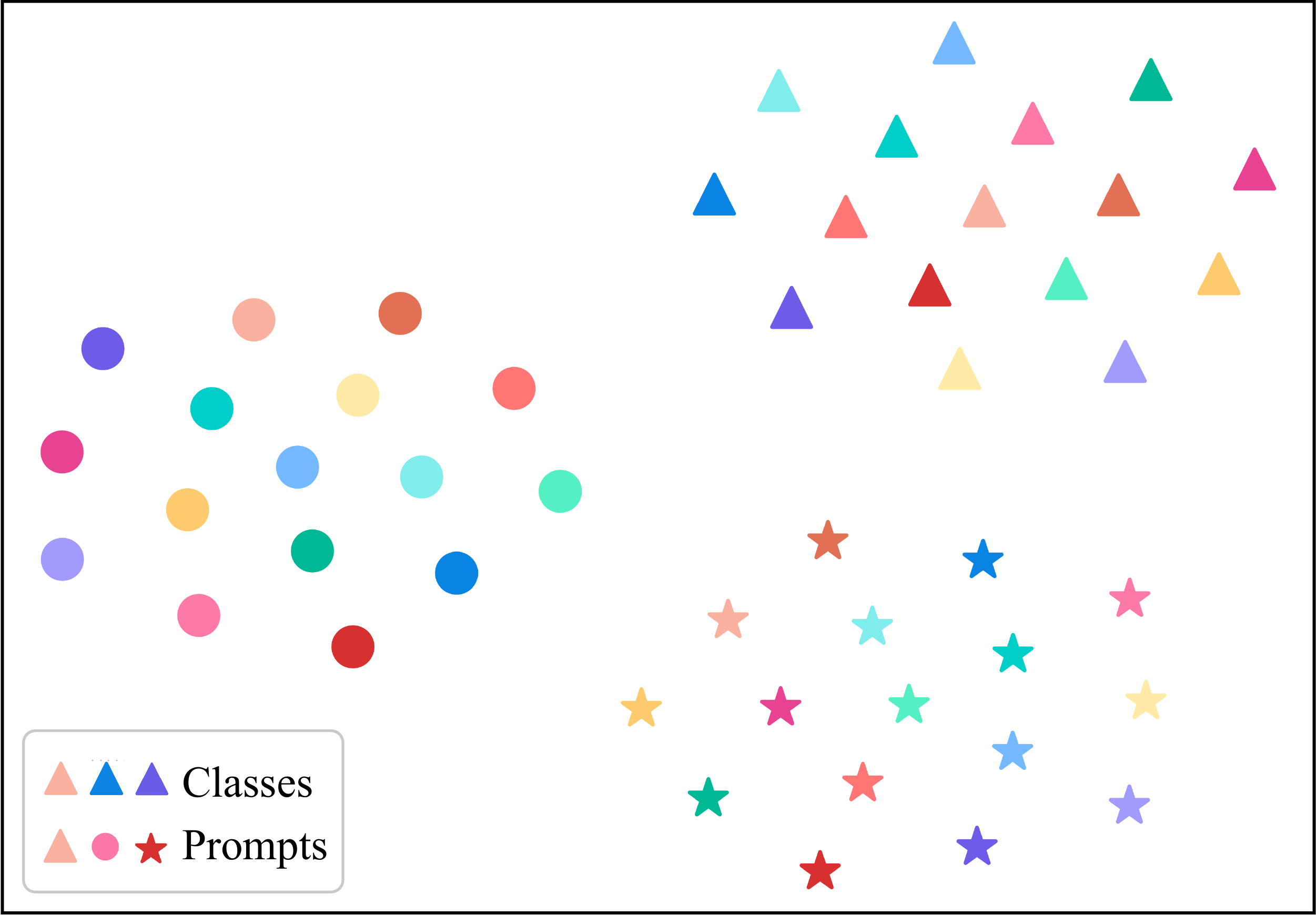}
\end{center}
   \caption{The illustration of text embedding of different prompts with different categories. Different colors represent different categories, and different shapes represent different prompts. Best viewed in color. }
\label{fig:promptvis}
\end{figure}

\subsection{Ablation Studies on MVP-SEG}

\noindent \textbf{Comparison with baseline}.
We adopt MaskCLIP~\cite{maskclip} as our baseline. MaskCLIP feeds handcrafted prompts into the CLIP text encoder with 85 templates as described in~\cite{gu2021open}. 

We compare segmentation performance with MaskCLIP on the unseen classes to demonstrate the zero shot ability of MVP-SEG. As show in Tab.~\ref{tab:ablation}, MVP-SEG achieves significant improvements (+8.4\% in COCO Stuff, +12.7\% in PASCAL VOC) on the unseen classes over baseline method using only 4 learned multi-view prompts (1 global classification prompt and 3 multi-view segmentation prompts). This result shows that, multi-view learnable prompts effectively contribute to the improvement of CLIP performance on pixel-level tasks.



\begin{figure*}[t]
\begin{center}
\includegraphics[width=.98\linewidth]{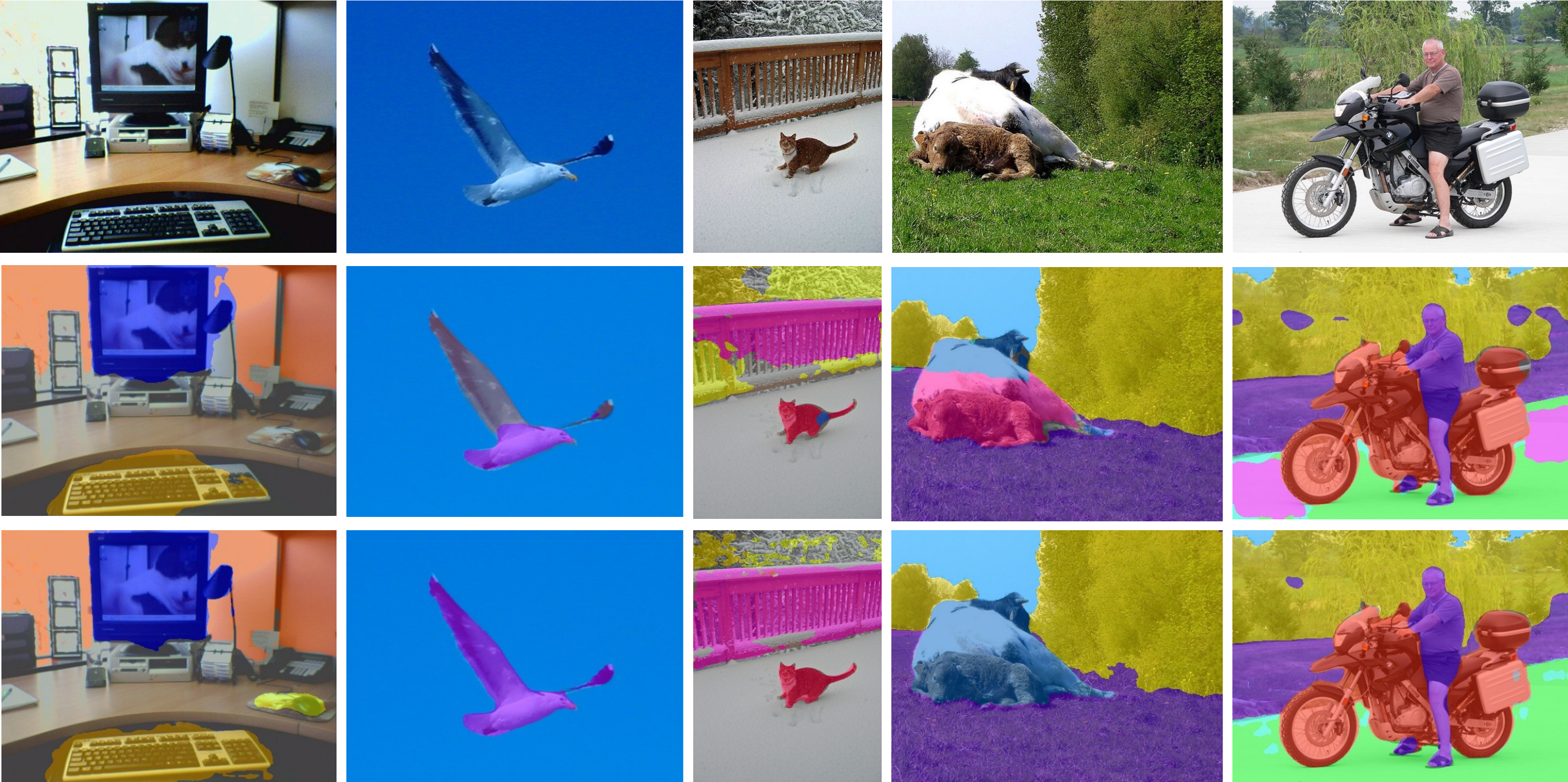}
\end{center}
   \caption{Comparison of segmentation results between MVP-SEG+ and MaskCLIP+~\cite{maskclip}. The first row is the input image, the second row is the prediction result of MaskCLIP+ and the third row is our result.}
\label{fig:mvp_mask_compare}
\end{figure*}

\begin{table*}[t]
  \centering
  \begin{tabular}{l | c | c | c }
    \toprule
    \textbf{} 
    mIoU(U)
    & Pascal VOC (20)
    & Pascal Context (59)
    & COCO Stuff (171)
    \\
    \midrule
    baseline                 &40.2 &26.9 &12.2 \\
    Pascal VOC (20)          &45.9 &29.8 &13.9 \\
    Pascal Context (59)      &47.8 &35.7 &19.5 \\
    COCO Stuff (171)         &\textbf{50.0} &\textbf{38.7} &\textbf{19.9} \\
    \bottomrule
  \end{tabular}
  \caption{Cross-dataset generalization test experiments. Columns represent different test sets, and rows represent the dataset used to learn prompts. Baseline indicates default prompts in MaskCLIP.} 
  \vspace{-0.6cm}
  \label{tab:generaize}
\end{table*}

\noindent \textbf{Study on segmentation prompts}.
As shown in Tab.~\ref{tab:ablation}, by replacing 85 manual prompts (as shown in the first row) with 1 learnable prompt (as in row 2), we get 3.6\% performance improvement on unseen classes on the COCO Stuff dataset and 1.6\% improvement on PASCAL VOC dataset. This result reveals the efficacy of learnable prompts for adapting CLIP features for segmentation. Increasing the number of learnable prompts is also beneficial for the performance. After deploying the OCLoss (as shown in row 4), the mIoU on unseen classes on COCO Stuff continues to increase from 16.4\% to 19.9\%, further indicating that our insight on multi-view prompt learning method is effective. We further test the optimal number of prompts with experiments on the COCO Stuff dataset. As depicted in Fig.~\ref{fig:num_prompts}, the performance (on unseen classes using OCLoss) increases as the number of prompts increases and reaches the peak at 3 prompts, thus our default number for multi-view segmentation prompts is 3.


We also visualize text embeddings under different learnable prompts in MVP-SEG with t-SNE~\cite{van2008visualizing}. As illustrated in Fig.~\ref{fig:promptvis}, text embeddings of different prompts are scattered in the feature space, which proves that the learned multi-view segmentation prompts are diverse. Moreover, visualization in Fig.~\ref{fig:fig1} shows that the learned prompts do contain high-level reasonable semantic information. More importantly, the segmentation results using different prompts tend to be complementary such that by combining them together can yield more accurate and complete masks. For more visualizations please see the Appendix.

To further verify the advantages of multi-view learnable prompts, we also compare the segmentation performance of our prompts with manually designed ones. We select a set of animal-related classes including \textit{cat, dog, horse, cow, bear, giraffe} 
 from COCO Stuff for comparison. As stated in Sec.~\ref{sec:intro}, We carefully hand-pick prompts related to body parts (\textit{ ``the leg of \{\}'',``the head of \{\}'', and `` the tail of \{\}'' }) for animal-related segmentation. As shown in Tab.~\ref{tab:tab5}, manual prompts outperform the MaskCLIP baseline with 1.0\% mIoU, indicating the feasibility of adopting multiple handcrafted prompts for segmentation. More importantly, we outperform both MaskCLIP and handcrafted prompts, showing the superiority of MVP-SEG.

\begin{table*}[t]
\centering
\footnotesize
\setlength\tabcolsep{5pt}
\scalebox{1}{
\begin{tabular}{c c c c c c c c c c c c c c}
\toprule
\multirow{2}{*}{Method}
&\multirow{2}{*}{ST}
&\multicolumn{4}{c}{COCO Stuff}
&\multicolumn{4}{c}{Pascal VOC}
&\multicolumn{4}{c}{Pascal Context}
\\
\cmidrule(lr){3-6}
\cmidrule(lr){7-10}
\cmidrule(lr){11-14}
&
&mIoU(U)
&mIoU(S)
&mIoU
&hIoU
&mIoU(U)
&mIoU(S)
&mIoU
&hIoU
&mIoU(U)
&mIoU(S)
&mIoU
&hIoU
\\
\midrule
SPNet~\cite{spnet}     & \ding{55} & 8.7     & 35.2    & 32.8    & 14.0    & 15.6    & 78.0    & 63.2    & 26.1    & -       & -       & -       & -    \\
ZS3Net~\cite{zs3net}    & \ding{55} & 9.5     & 34.7    & 33.3    & 15.0    & 17.7    & 77.3    & 61.6    & 28.7    & 12.7    & 20.8    & 19.4    & 15.8 \\ 
CaGNet~\cite{cagnet}    & \ding{55} & 12.2    & 35.5    & 33.5    & 18.2    & 26.6    & 78.4    & 65.5    & 39.7    & 18.5    & 24.8    & 23.2    & 21.2 \\ 
SIGN~\cite{sign}      & \ding{55} & 15.5    & 32.3    & -       & 20.9    & 28.9    & 75.4    & -       & 41.7    & -       & -       & -       & -    \\
Joint~\cite{joint}     & \ding{55} & -       & -       & -       & -       & 32.5    & 77.7    & -       & 45.9    & 14.9    & 33.0    & -       & 20.5 \\ 
ZegFormer~\cite{zegformer} & \ding{55} & 33.2    & 36.6    & -       & 34.8    & 63.6    & 86.4    & -       & 73.3    & -       & -       & -       & -    \\
\midrule
\midrule
SPNet~\cite{spnet}                     &\ding{51}  & 26.9    & 34.6    & 34.0    & 30.3    & 25.8    & 77.8    & 64.8    & 38.8    & -       & -       & -       & -    \\
ZS3Net~\cite{zs3net}                    &\ding{51}  & 10.6    & 34.9    & 33.7    & 16.2    & 21.2    & 78.0    & 63.0    & 33.3    & 20.7    & 27.0    & 26.0    & 23.4 \\ 
CaGNet~\cite{cagnet}                    &\ding{51}  & 13.4    & 35.5    & 33.7    & 19.5    & 30.3    & 78.6    & 65.8    & 43.7    & -       & -       & -       & -    \\
SIGN~\cite{sign}                      &\ding{51}  & 17.5    & 31.9    & -       & 22.6    & 41.3    & 83.5    & -       & 55.3    & -       & -       & -       & -    \\
STRICT~\cite{strict}                    &\ding{55}  & 30.3    & 35.3    & 34.9    & 32.6    & 35.6    & 82.7    & 70.9    & 49.8    & -       & -       & -       & -    \\
MaskCLIP+~\cite{maskclip}                 &\ding{55}  & 54.7    & -       & 39.6    & 45.0    & 86.1    & -       & 88.1    & 87.4    & 66.7    & -       & 48.1    & 53.3 \\ 
\midrule
\midrule
MVP-SEG+                  &\ding{51}  & \textbf{55.8}    & \textbf{38.3}    & \textbf{39.9}    & \textbf{45.5}    & \textbf{87.4}    & \textbf{89.0}    & \textbf{88.6}    & \textbf{88.2}    & \textbf{67.5}    & \textbf{44.9}    & \textbf{48.7}    & \textbf{54.0} \\ 
\midrule
Fully Sup~\cite{maskclip}                 &\ding{55}  & -       & -       & \textbf{39.9}    & -       & -       & -       & 88.2    & -       & -       & -       & 48.2    & -   \\ 

\bottomrule
\end{tabular}

}

\caption{Comparison with the SOTA methods. Depending on whether the unseen
classes are visible during training, previous methods can be divided into inductive methods(top part) and transductive methods(middle part). Our method belongs to the latter. MVP-SEG+ and Fully Sup indicate the DeepLab models train by MVP-SEG pseudo label fully-supervised annotations respectively.}
\label{tab:compare_with_sota}
\end{table*}
 
 \begin{figure*}[t]
 \begin{center}
 \includegraphics[width=.98\linewidth]{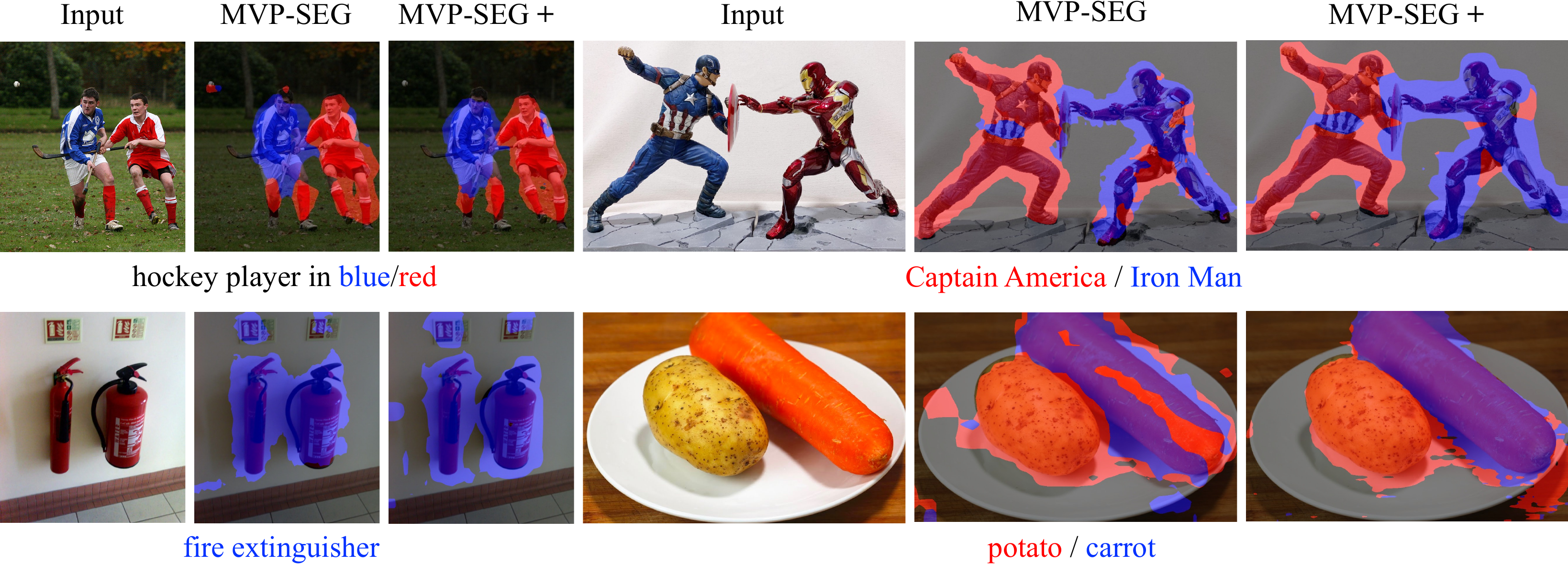}
 \end{center}
    \caption{Qualitative results of MVP-SEG and MVP-SEG+ on open-vocabulary web images.}
 \label{fig:arbitary}

 \end{figure*}

\noindent \textbf{The influence of GPR}.
We use the GPR module to further refine segmentation masks with the image-level classification capability of CLIP. Row 5 of Tabel~\ref{tab:ablation} shows that adding GPR can comprehensively improve performances on both seen and unseen classes. hIoU result on COCO Stuff dataset is improved from 17.9\% to 19.2\%, and on PASCAL VOC increases from 48.3\% to 53.1\%. Visualization results also show that GPR can improve segmentation by filtering out masks of false positive classes (please refer to the Appendix).

\noindent \textbf{The generalization of multi-view learnable prompts}. Experiment results in Tab.~\ref{tab:ablation} show that learned prompts not only improve performances on seen classes but also significantly boost unseen classes as well. To further study the generalization ability of our learned prompts, we evaluate the performance by training prompts on one dataset but test on other datasets. As shown in Tab.~\ref{tab:generaize}, prompts trained on other datasets outperform the baseline by a large margin. 


\subsection{Comparison with State-of-the-Art}

From Tab.\ref{tab:compare_with_sota}, we can conclude: 1) Our method can effectively improve the performance of novel classes. MVP-SEG+ improves the previous SOTA method by 1.1\%, 1.3\% and 0.8\% mIoU of unseen classes on COCO Stuff, Pascal VOC and Pascal Context respectively; 2) Our method consistently outperforms previous SOTA on all datasets(+0.3\%,+0.5\% and +0.6\% hIoU); 3) Our proposed method can effectively adapt powerful CLIP to pixel level tasks so that the performance of MVP-SEG+ is competitive or even surpass fully supervised counterparts. To the best of our knowledge, this is the first time such a result has been achieved.

Visual comparison with MaskCLIP+, which is one of the SOTAs, is illustrated in Fig.~\ref{fig:mvp_mask_compare}, and our method obtains more complete masks compared, and false positive classes can be filtered. Furthermore, we also illustrate the performance of our method on web-crawled images which contain rare object classes that are not visible in any training set. As shown in Fig.~\ref{fig:arbitary}, MVP-SEG can correctly attend to rare objects such as Iron Man and Captain America, and MVP-SEG+ segments them with favorable accuracy. 



\section{Conclusion}
In this work, we propose multi-view prompt learning (MVP-SEG) to settle open-vocabulary semantic segmentation with pre-trained CLIP. At first, we demonstrate the efficacy of adapting pre-trained CLIP model from image-to-pixel level for open-vocabulary segmentation, then introduce multi-view prompt learning, inspired by part-based representation, to convey this adaptation. Multi-view learnable prompts are optimized by our Orthogonal Constraint Loss (OCLoss) to ensure the part-wise attention of each prompt so that collaboratively they yield better segmentation results. In addition, we design Global Prompt Refining (GPR) adopting a global learnable prompt to remove class-wise noises and refine segmentation masks. MVP-SEG+ reports SOTA open-vocabulary segmentation performance on all three widely-used benchmarks and reveals superior results than fully-supervised counterparts on PASCAL VOC and PASCAL Context datasets.

{\small
\bibliographystyle{ieee_fullname}
\bibliography{egbib}
}

\end{document}